\pdfoutput=1

\documentclass[10pt,twocolumn,letterpaper]{article}

\usepackage[pagenumbers]{cvpr} %

\usepackage{multirow}
\usepackage{colortbl}
\usepackage{tikz}
\usetikzlibrary{positioning,arrows.meta,fit,backgrounds}
\newcommand{\method}{CANDOR\xspace}

\definecolor{oursrow}{rgb}{0.93,0.96,1.0}
\newlength{\qualw}
\newlength{\qbw}

\definecolor{cvprblue}{rgb}{0.21,0.49,0.74}
\usepackage[pagebackref,breaklinks,colorlinks,allcolors=cvprblue]{hyperref}

\def\confName{CVPR}
\def\confYear{2027}

\title{CANDOR: Contrastive Self-Denoising Rewards for Label-Free Text-to-Image~Post-Training}

\author{Junhyeok Lee$^{1}$\qquad Byung-Hoon Kim$^{2}$\qquad Kyu Sung Choi$^{1}$\\
$^{1}$Seoul National University\qquad $^{2}$EverEx}

\begin{document}
\twocolumn[{%
\renewcommand\twocolumn[1][]{#1}%
\maketitle
\begin{center}
\vspace{-1.5em}
\setlength{\qualw}{0.88\textwidth}%
\setlength{\tabcolsep}{0pt}%
\begin{tabular}{@{}c@{\hspace{2pt}}c@{}}
\rotatebox{90}{\makebox[0.1975\qualw][c]{\small Base model}} & \includegraphics[width=\qualw]{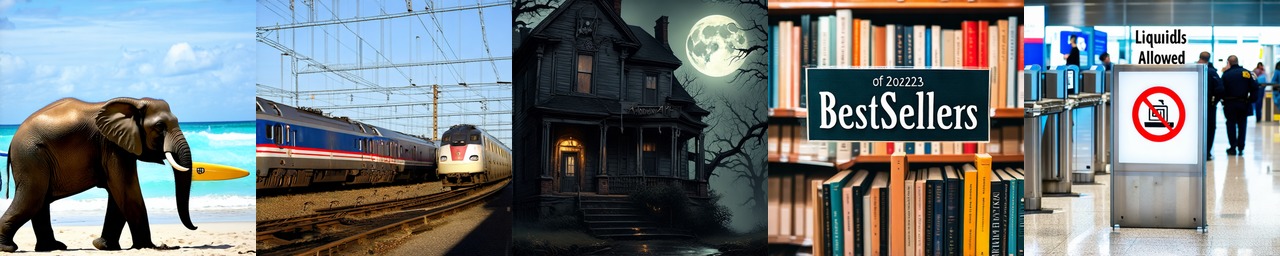} \\[-\dp\strutbox]
\rotatebox{90}{\makebox[0.1975\qualw][c]{\small + \method}} & \includegraphics[width=\qualw]{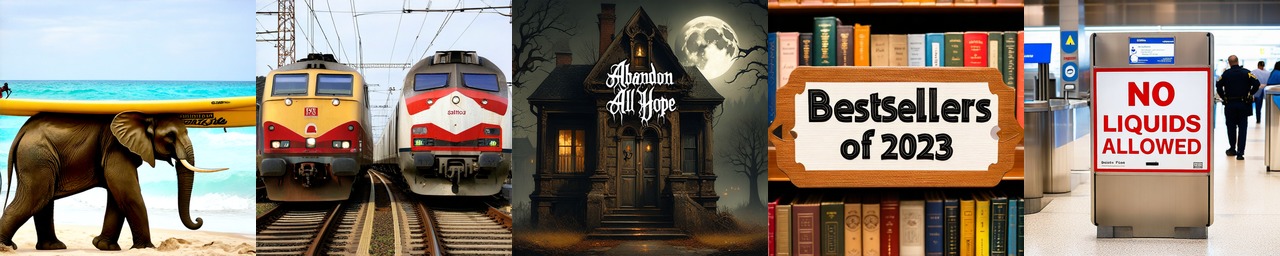} \\
& {\setlength{\tabcolsep}{0.0075\qualw}\setlength{\qualw}{0.925\qualw}\begin{tabular}{ccccc}\parbox[t]{0.2\qualw}{\centering\footnotesize a photo of an elephant below a surfboard} & \parbox[t]{0.2\qualw}{\centering\footnotesize a photo of two trains} & \parbox[t]{0.2\qualw}{\centering\footnotesize ``Abandon All Hope''} & \parbox[t]{0.2\qualw}{\centering\footnotesize ``Bestsellers of 2023''} & \parbox[t]{0.2\qualw}{\centering\footnotesize ``No Liquids Allowed''}
\end{tabular}}
\end{tabular}
\vspace{-0.5em}
\captionof{figure}{\textbf{Base model before and after \method post-training.} GenEval (left) and OCR text-rendering (right) test prompts; OCR labels: text to render. Hand-picked to show differences.}
\label{fig:teaser}
\vspace{0.5em}
\end{center}
}]

\begin{abstract}
Post-training text-to-image generators with reinforcement learning usually relies on external reward models that need annotated preferences or auxiliary evaluators and are prone to reward hacking. We show that the generator's own denoising accuracy, used by a recent intrinsic reward, is a poor proxy for prompt fidelity. Within a same-prompt group, this absolute self-confidence is dominated by image complexity and ranks correct samples above incorrect ones near chance. We introduce \method (\textbf{C}ontrastive \textbf{A}lignment from \textbf{N}ull-referenced \textbf{D}enoising l\textbf{O}g-\textbf{R}atio), the log-ratio of unconditional to conditional denoising errors under shared noise probes, which measures how much the prompt helps the model denoise its own sample. It cancels prompt-independent difficulty that scales both errors and ranks the better of two same-prompt samples higher in $60\%$ and $63\%$ of pairs on compositional generation and text rendering. As the only Flow-GRPO reward for a flow-matching model, \method improves text rendering, aesthetic and human-preference scores over the base model and a prior absolute self-confidence reward, with text-rendering and preference gains that stronger classifier-free guidance does not reproduce. On public Flow-GRPO adapters trained on external rewards, \method further raises PickScore and ImageReward, so it can complement those rewards. Code is available at \url{https://github.com/jhlee0619/candor}.
\end{abstract}

\section{Introduction}
\label{sec:intro}

Text-to-image (T2I) generators~\cite{rombach2022ldm,saharia2022imagen,esser2024sd3} produce high-fidelity images, yet often miss requested objects, spatial relations or text~\cite{ghosh2023geneval,liu2025flowgrpo}. Reinforcement-learning post-training narrows this gap by updating the generator toward highly rewarded samples~\cite{black2023ddpo,fan2023dpok,liu2025flowgrpo}.

These rewards are almost always \emph{external}: human-preference models~\cite{kirstain2023pickscore,wu2023hpsv2,xu2023imagereward} or validators such as object detectors and optical character recognition (OCR) engines~\cite{ghosh2023geneval,cui2025paddleocr}. They need annotations or auxiliary networks, and overoptimizing a learned proxy degrades the objective it stands for~\cite{gao2023scaling}: T2I policies exploit external rewards with high-scoring but artifact-prone images~\cite{zhang2024confronting,hong2026understanding,liu2025flowgrpo}. \emph{Intrinsic} rewards remove this dependence by letting the generator judge its own samples, in language models~\cite{yuan2024selfrewarding,zhao2025intuitor,agarwal2025entropy,zuo2025ttrl} and image generators~\cite{chen2025iris,pan2026gvu}. For a diffusion or flow model, how accurately it recovers noise injected into one of its own samples is a self-confidence score, and rewarding this \emph{absolute} self-confidence improves T2I models~\cite{kim2026solace}. This reward implicitly assumes that samples the model denoises well follow the prompt.

This assumption breaks: a sample can also be easy to denoise because it is \emph{simple}, as with a flat background or a texture-free object. This simplicity confound matters most for group relative policy optimization (GRPO)~\cite{shao2024grpo} as used by Flow-GRPO~\cite{liu2025flowgrpo}. GRPO centers rewards within each prompt's group, so only their relative values drive the update, and per-group scaling amplifies small differences~\cite{wang2025prefgrpo}. Within same-prompt groups, absolute self-confidence is dominated by image complexity and ranks correct samples above incorrect ones near chance, so it can steer the policy toward simpler rather than more faithful images. A generator should instead reward itself for what the prompt explains.

Image complexity makes a sample harder to denoise \emph{with or without} the prompt, whereas prompt fidelity should only help the prompt-conditioned prediction. We therefore reward the \emph{contrast} between the conditional and unconditional predictions, on the same re-noised latent and noise:
\begin{equation}
D = \log \mathrm{MSE}(\varnothing) - \log \mathrm{MSE}(c),
\label{eq:d_intro}
\end{equation}
where $\mathrm{MSE}(\varnothing)$ and $\mathrm{MSE}(c)$ are the mean squared errors (MSE) of the model's noise estimate without and with the prompt. A log-ratio cancels any difficulty factor that scales both errors, so $D$ is large when the prompt explains the sample. Models that support classifier-free guidance (CFG)~\cite{ho2022cfg} already have an unconditional (empty-prompt) branch, so $D$ needs no additional model. Prior work uses this contrast to evaluate or interpret fixed models~\cite{krojer2023diffusionitm,kong2024interpretable} or to select samples for supervised fine-tuning~\cite{wang2025mitune,wang2025rfmi}. We use it as an online GRPO reward on the policy's own samples and call it \method (\textbf{C}ontrastive \textbf{A}lignment from \textbf{N}ull-referenced \textbf{D}enoising l\textbf{O}g-\textbf{R}atio).

The lesson for intrinsic rewards is that self-denoising should be read contrastively: what matters is how much the prompt helps the model denoise a sample, not how well the model denoises it. Read this way, self-denoising can also complement external rewards: \method further raises the PickScore and ImageReward of public Flow-GRPO adapters.

Our contributions are:
\begin{itemize}
\item We show that, within the same-prompt groups of GRPO, absolute self-confidence is dominated by image simplicity rather than prompt fidelity.
\item We propose \method, a label-free intrinsic reward that measures how much the prompt helps the model denoise its own samples. Within same-prompt groups on compositional generation and text rendering, it is nearly uncorrelated with image complexity and ranks the better sample of a pair higher in $60\%$ and $63\%$ of pairs, whereas absolute self-confidence is near chance.
\item As the only GRPO reward, \method improves text rendering, aesthetics and human preference over the base model and absolute self-confidence on external metrics that the reward never sees (\cref{fig:teaser}, Tabs.~\ref{tab:main}--\ref{tab:cfg}), beyond what stronger guidance achieves.
\end{itemize}

\section{Related work}
\label{sec:related}

\paragraph{Post-training with external rewards.}
T2I diffusion~\cite{ho2020ddpm,rombach2022ldm,saharia2022imagen} and flow-matching models~\cite{lipman2023flowmatching,liu2023rectifiedflow,esser2024sd3} have been aligned through differentiable rewards~\cite{xu2023imagereward}, policy gradients~\cite{black2023ddpo,fan2023dpok} and preference optimization~\cite{wallace2024diffusiondpo}. Flow-GRPO~\cite{liu2025flowgrpo} applies GRPO~\cite{shao2024grpo} to flow matching by turning the deterministic sampler into a marginal-preserving stochastic differential equation (SDE). Overoptimizing external rewards yields artifact-prone images~\cite{zhang2024confronting}, which ensembling only partly mitigates~\cite{hong2026understanding}. We keep Flow-GRPO but compute the reward from the policy itself.

\paragraph{Intrinsic rewards.}
IRIS~\cite{chen2025iris} trains autoregressive T2I models to \emph{minimize} token-level self-certainty, after observing that highly certain models produce simple, uniform images. For diffusion T2I models, SOLACE~\cite{kim2026solace} rewards absolute self-confidence and reports gains in compositional generation and text rendering. We study what this absolute signal measures within prompt groups, show that it is dominated by image simplicity, and replace it with a contrastive log-ratio against the unconditional prediction.

\paragraph{Likelihood ratios and input complexity.}
Deep generative models can assign high likelihood to simple out-of-distribution inputs~\cite{nalisnick2019know}, partly through local pixel correlations that transfer across datasets~\cite{kirichenko2020normalizing}. For out-of-distribution detection, likelihood ratios against a background model~\cite{ren2019likelihood} and compression-based corrections~\cite{serra2020input} remove this input-complexity effect. In diffusion models, complexity affects the low-noise evidence-lower-bound terms most, and a ratio of low-noise to total terms corrects for it~\cite{goodier2023likelihood}. Self-denoising confidence as a reward inherits this confound. We measure it \emph{within} same-prompt groups, where GRPO consumes rewards, and correct it with the model's own unconditional prediction rather than a separate background model or a ratio across noise levels.

\paragraph{Diffusion models as implicit classifiers.}
CFG~\cite{ho2022cfg} applies the gradient of an implicit classifier $\log p(c\,|\,x_t)$, the difference between conditional and unconditional predictions, at sampling time. Diffusion classifiers compare denoising errors across conditions for zero-shot recognition~\cite{li2023diffusionclassifier,clark2023texttoimage}. For fixed models, DiffusionITM~\cite{krojer2023diffusionitm} normalizes the conditional error by the unconditional one for image--text matching, and Kong et al.~\cite{kong2024interpretable} estimate image--text pointwise mutual information from denoising errors. MI-Tune~\cite{wang2025mitune} scores samples by the conditional--unconditional noise-prediction gap, keeps the best of 50 per prompt, and fine-tunes on them with the supervised diffusion loss. RFMI~\cite{wang2025rfmi} does the same for rectified-flow models, including the base model we post-train. SoftREPA~\cite{lee2025softrepa} fine-tunes soft text tokens contrastively on matched and mismatched image--text pairs. MIPO~\cite{nam2026mipo} maximizes prompt--response pointwise mutual information in language models, and guidance-free training~\cite{chen2025cca,chen2025gft} makes sampling without CFG match guided sampling. \method uses the log-ratio of the unconditional to the conditional error as an online GRPO reward on the policy's own samples.

\section{Preliminaries}
\label{sec:prelim}

\subsection{GRPO for flow matching}
Rectified flow~\cite{liu2023rectifiedflow} interpolates $x_t=(1-t)\,x_0+t\,x_1$ between data $x_0$ and noise $x_1\!\sim\!\mathcal{N}(0,I)$ and trains a network $v_\theta(x_t,t,c)$ to regress the constant velocity $x_1-x_0$. Flow-GRPO~\cite{liu2025flowgrpo} samples a group of $G$ trajectories (rollouts) per prompt $c$ with an SDE sampler. It centers the rewards of the terminal latents $z_0^{(i)}$ within the group and, in its implementation, scales them by the batch-wide standard deviation,
\begin{equation}
\hat{A}^i=\frac{R(z_0^{(i)},c)-\mathrm{mean}(\{R(z_0^{(j)},c)\}_{j=1}^G)}{\mathrm{std}(\{R\}_{\mathrm{batch}})+10^{-4}}.
\label{eq:adv}
\end{equation}
It then updates the policy $\pi_\theta$ (the SDE sampler built on $v_\theta$) by maximizing a clipped importance-ratio objective with a Kullback--Leibler (KL) penalty toward the reference model. \Cref{eq:adv} shows the property that guides our design. The numerator compares each sample only with the others in its prompt group, so a reward matters only through how it compares samples of the same prompt. The batch-wide denominator sets the scale of the update.

\subsection{Self-denoising confidence}
Let $z_0^{(i)}$ be the terminal latent of the $i$-th sample. We draw $K$ noise probes $\epsilon^{(i,m)}\!\sim\!\mathcal{N}(0,I)$ independently for each sample, paired antithetically ($\epsilon^{(i,m+K/2)}=-\epsilon^{(i,m)}$), and re-noise the latent at a set $\mathcal{T}$ of the sampler's timesteps:
\begin{equation}
z_t^{(i,m)}=(1-t)\,z_0^{(i)}+t\,\epsilon^{(i,m)}.
\label{eq:renoise}
\end{equation}
Under the rectified-flow parameterization, the noise implied by the velocity prediction at the known terminal latent $z_0$ is $\hat\epsilon_\theta(z_t,t,y)=v_\theta(z_t,t,y)+z_0$ for a condition $y$, and the self-denoising MSE is
\begin{equation}
\mathrm{MSE}_{i,t}(y)=\frac{1}{Kd}\sum_{m=1}^K\big\|\hat\epsilon_\theta(z_t^{(i,m)},t,y)-\epsilon^{(i,m)}\big\|_2^2,
\label{eq:mse}
\end{equation}
where $d$ is the number of latent elements. The \emph{absolute self-confidence} of a sample is $S_{i,t}=-\log(\mathrm{MSE}_{i,t}(c)+\delta)$ with $\delta\!=\!10^{-6}$. Normalized across the batch at each timestep and averaged over $t$, it serves as the intrinsic reward of SOLACE~\cite{kim2026solace}. Errors are computed without CFG and directly on the rollout latents, because re-encoding the decoded image with the autoencoder would change the low-noise errors substantially.

\section{Method: \method}
\label{sec:method}

\begin{figure*}[t]
\centering
\resizebox{\textwidth}{!}{%
\begin{tikzpicture}[
  font=\footnotesize, node distance=0.4cm,
  box/.style={draw, line width=0.8pt, rounded corners=2pt, align=center, inner sep=3pt, fill=white},
  model/.style={box, draw=black!45, fill=black!4, minimum width=0.8cm, minimum height=2.9cm},
  cond/.style={box, draw=orange!80!black, fill=orange!12, minimum width=2.0cm, minimum height=0.75cm},
  unc/.style={box, draw=gray!70!black, fill=gray!12, minimum width=2.0cm, minimum height=0.75cm},
  thumb/.style={inner sep=0pt, draw=black!60, line width=0.6pt},
  item/.style={box, draw=black!60, fill=white, minimum width=1.05cm, minimum height=0.42cm, inner sep=1pt},
  panel/.style={draw, dashed, line width=0.6pt, rounded corners=6pt, inner sep=5pt},
  grpbox/.style={draw=black!45, dashed, line width=0.5pt, rounded corners=3pt, inner sep=3pt},
  arr/.style={-{Stealth[length=4pt]}, line width=0.8pt, draw=black}]
\begin{scope}[color=black!55, every node/.append style={text=black}, font=\normalsize, shift={(1.32,-0.31)}, scale=0.8, transform shape]
\node[model, minimum height=2.3cm] (gen) at (1.65,0.3) {\rotatebox{90}{\shortstack{T2I model $\pi_\theta$\\(SDE)}}};
\node[box, minimum width=1.5cm] (c) at (1.65,-1.6) {Text prompt $c$};
\draw[arr] (c.north) -- (gen.south);
\foreach \k in {2,1,0} \node[thumb] (z0\k) at (4.0+0.09*\k,0.25+0.09*\k) {\includegraphics[width=1.25cm]{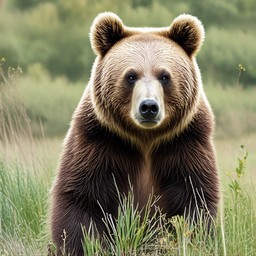}};
\draw[arr, dashed] (gen.east |- z00) -- node[above]{sample} (z00.west);
\node[box, anchor=south east] (eps) at (7.58,1.75) {\begin{tabular}{@{}c@{}}antithetic pair\\$+\epsilon^{(m)}$\;\;$-\epsilon^{(m)}$\end{tabular}\;$\vcenter{\hbox{\includegraphics[width=0.7cm]{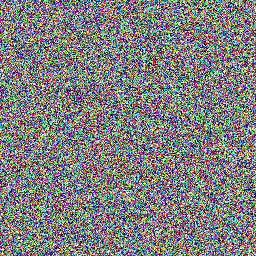}\,\includegraphics[width=0.7cm]{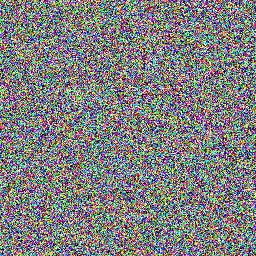}}}$};
\begin{scope}[shift={(5.7,-0.55)}]
\node[thumb, anchor=south west] (g11) at (0,0.98) {\includegraphics[width=0.9cm]{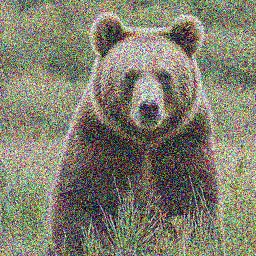}};
\node[thumb, anchor=south west] (g12) at (0.98,0.98) {\includegraphics[width=0.9cm]{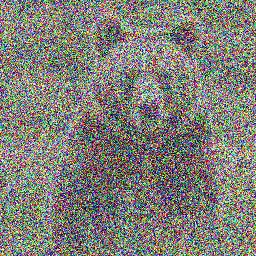}};
\node[thumb, anchor=south west] (g21) at (0,0) {\includegraphics[width=0.9cm]{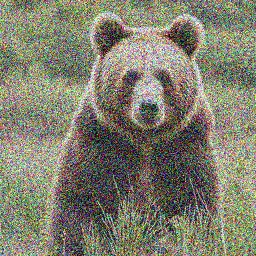}};
\node[thumb, anchor=south west] (g22) at (0.98,0) {\includegraphics[width=0.9cm]{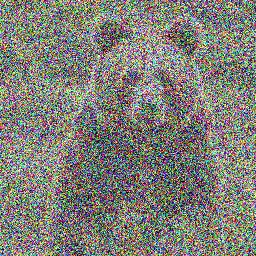}};
\draw[-{Stealth[length=3pt]}] (-0.12,-0.12) -- (2.1,-0.12) node[right]{$t$};
\draw[-{Stealth[length=3pt]}] (-0.12,-0.12) -- (-0.12,2.0) node[left]{$\pm$};
\end{scope}
\node[align=center, below=0.3cm of g21.south east] (ztl) {Re-noised latents\\$\{z_t^{(i,m)}\}$};
\node[align=center] (z0l) at (z00 |- ztl) {Terminal latents\\$\{z_0^{(i)}\}_{i=1}^G$};
\draw[arr] (z00.east) -- ++(0.35,0) |- ([xshift=-0.12cm]g21.west);
\draw[arr] (z00.east) -- ++(0.35,0) |- ([xshift=-0.12cm]g11.west);
\draw[arr] (6.64,0 |- eps.south) -- ([yshift=0.05cm]6.64,0 |- g11.north);
\end{scope}
\node[cond] (vc) at (9.6,1.2) {$\hat\epsilon_\theta(z_t^{(i,m)},t,c)$};
\node[unc] (vu) at (9.6,-1.2) {$\hat\epsilon_\theta(z_t^{(i,m)},t,\varnothing)$};
\draw[line width=0.8pt] (g12.east) -| (7.65,0) |- (g22.east);
\draw[arr] (7.65,0) -- (8.0,0) |- (vc.west);
\draw[arr] (8.0,0) |- (vu.west);
\node[thumb, draw=orange!80!black] (gc) at (12.0,1.2) {\includegraphics[width=1.0cm]{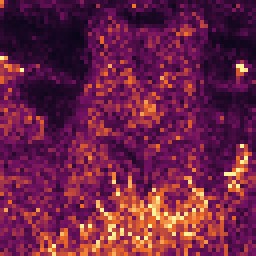}};
\node[thumb] (gu) at (12.0,-1.2) {\includegraphics[width=1.0cm]{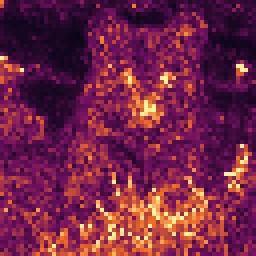}};
\node[right=2pt of gc, align=left] (lc) {\phantom{$\mathrm{MSE}(c)$}\\\phantom{$=0.225$}};
\node[right=2pt of gu, align=left] (lu) {\phantom{$\mathrm{MSE}(\varnothing)$}\\\phantom{$=0.236$}};
\draw[arr] (vc.east) -- (gc.west);
\draw[arr] (vu.east) -- (gu.west);
\node[box, draw=orange!80!black, fill=orange!6, inner sep=3pt] (d) at (12.0,0) {$\displaystyle D_{i,t}=\log\frac{\mathrm{MSE}(\varnothing)+\delta}{\mathrm{MSE}(c)+\delta}$};
\draw[arr] (gc.south) -- (d.north);
\draw[arr] (gu.north) -- (d.south);
\foreach \n/\y in {1/1.1, 2/0.6, G/-1.1} {
  \node[item] (r\n) at (15.3,\y) {$R_{\n}$};
  \node[item] (a\n) at (18.1,\y) {$\hat A^{\n}$};}
\node at (15.3,-0.2) {$\vdots$}; \node at (18.1,-0.2) {$\vdots$};
\node (rh) at (15.3,1.65) {rewards}; \node (ah) at (18.1,1.65) {advantages};
\node[grpbox, fit=(r1)(rG)] (rbox) {}; \node[grpbox, fit=(a1)(aG)] (abox) {};
\node[box, minimum height=0.9cm] (grp) at (16.7,0) {Group\\norm.};
\draw[arr] (d.east) -- (rbox.west |- d.east);
\draw[arr] (rbox.east |- grp.west) -- (grp.west); \draw[arr] (grp.east) -- (abox.west |- grp.east);
\coordinate (top) at (0,2.15); \coordinate (bot) at (0,-1.95);
\begin{scope}[on background layer]
\node[panel, draw=black!20, fill=black!1, fit=(c)(gen)(z0l)(eps)(ztl)(c|-top)(c|-bot), label={[font=\footnotesize]above:Rollout \& re-noising}] (pa) {};
\node[panel, draw=black!35, fill=white, fit=(vc)(vu)(d)(lc)(lu)(vc|-top)(vc|-bot), label={[font=\footnotesize]above:Contrastive denoising log-ratio}] (pb) {};
\node[panel, draw=black!35, fill=white, fit=(rbox)(abox)(grp)(rh|-top)(rh|-bot), label={[font=\footnotesize]above:GRPO}] (pc) {};
\end{scope}
\draw[arr, dashed] (pc.east) -- ++(0.25,0) |- (gen.north |- 0,2.95) node[pos=0.75, above]{\textbf{GRPO policy update}} -- (gen.north);
\end{tikzpicture}}
\vspace{-0.5em}
\caption{\textbf{Overview of \method.} Rollout \& re-noising: policy samples re-noised with antithetic probes. Contrastive denoising log-ratio: $D_{i,t}$ of empty-prompt to prompt errors (CFG off). GRPO: per-sample rewards (timestep mean of z-scored $D_{i,t}$) group-normalized into advantages. Error maps: latent error of this sample at $t{=}0.60$, averaged over probes and channels, one color scale; other thumbnails: illustrative pixel-space renderings.}
\label{fig:overview}
\vspace{-0.5em}
\end{figure*}

\subsection{Within-group analysis of absolute self-confidence}
\label{sec:analysis_offline}
\paragraph{Protocol.} Because GRPO compares rewards only within a prompt group (\cref{eq:adv}), we evaluate candidate rewards in that regime. For each test prompt of GenEval~\cite{ghosh2023geneval} and of the OCR text-rendering benchmark of Flow-GRPO~\cite{liu2025flowgrpo}, we draw $G\!=\!16$ base-model rollouts with the training SDE sampler at the evaluation CFG scale of $7.0$ and score them with the training reward probes. The official evaluator gives each image a correctness score $\ell$ (GenEval correctness or OCR edit-distance score). For a scalar score $s$, we report the \emph{within-group concordance} $\Pr[s(x_a)>s(x_b)\mid \ell_a>\ell_b]$ over all same-prompt pairs with different $\ell$. As a complexity proxy, we use the log PNG file size, a compression-based estimate as in~\cite{serra2020input}, z-scored within the group.

\paragraph{Correlation with complexity and correctness.} Within a group, absolute self-confidence $S$ correlates with complexity at $r\!=\!-0.75$ on GenEval and $-0.74$ on OCR (\cref{tab:offline}), so it ranks simple samples above complex ones. Its concordance with correctness is close to chance ($0.48$ on GenEval, $0.51$ on OCR).

\subsection{Contrastive denoising log-ratio reward}
\label{sec:candor}
Because image complexity makes a latent harder to denoise \emph{with or without} the prompt, we compare the conditional error with the unconditional error on the \emph{same} re-noised latents and probes (\cref{fig:overview}):
\begin{equation}
D_{i,t}=\log\big(\mathrm{MSE}_{i,t}(\varnothing)+\delta\big)-\log\big(\mathrm{MSE}_{i,t}(c)+\delta\big),
\label{eq:d}
\end{equation}
where $\varnothing$ is the empty (null) prompt, which the model already supports for CFG. A prompt-independent difficulty factor that scales both errors cancels in the ratio, and it cancels exactly when $\delta$ is negligible against both errors.

\paragraph{Interpretation.} For diffusion and flow models, the weighted denoising error bounds the negative log-likelihood up to constants~\cite{kingma2023elbo}. Heuristically, the error difference relates to $\log p(z_0\,|\,c)-\log p(z_0)=\log p(c\,|\,z_0)-\log p(c)$, the pointwise mutual information between sample and prompt. The difference of two bounds is neither a bound nor a guaranteed approximation of this quantity, so we use the connection as motivation only. Negating absolute self-confidence would simply reward complex samples. The log-ratio instead rewards a sample when the prompt makes it \emph{more} predictable than it would be unconditionally, however predictable it is in absolute terms.

\paragraph{Reward.} We z-score $D_{i,t}$ across the batch at each probed timestep and average over timesteps:
\begin{equation}
R_{\text{\method}}(z_0^{(i)},c)=\frac{1}{|\mathcal{T}|}\sum_{t\in\mathcal{T}}w(t)\,\frac{D_{i,t}-\mu_t}{\sigma_t},
\label{eq:reward}
\end{equation}
where $\mu_t$ and $\sigma_t$ are the batch mean and standard deviation of $D_{i,t}$ at timestep $t$, and $w(t)$ is a timestep weight ($w(t)\!=\!1$ in training). The reward enters GRPO through \cref{eq:adv}. Both errors use the current policy weights that also generate the rollouts, so there is no separate frozen scorer.

\subsection{Offline validation}
\label{sec:offline_val}
$D$ is nearly uncorrelated with both complexity and absolute self-confidence (\cref{tab:offline}). Of the three scores, only $D$ has a concordance clearly above chance on both benchmarks ($0.60$ on GenEval, $0.63$ on OCR), also with complexity regressed out. At the training CFG scale of 4.5, on training prompts disjoint from the evaluation prompts, $D$ reaches $0.615$ on both benchmarks, against $0.550$ and $0.522$ for $S$, which still correlates with complexity at $r\!=\!-0.72$ and $-0.65$ (supplementary material). Within a group, $S$ scores samples almost exactly like the unconditional confidence $-\log\mathrm{MSE}(\varnothing)$ (correlation $0.999$ on both benchmarks), so absolute self-confidence carries essentially no prompt-specific information.

\begin{table}[t]
\centering
\caption{\textbf{Offline analysis of intrinsic rewards.} Base-model test-prompt rollouts at CFG 7.0. Concordance: fraction of same-prompt pairs ranked correctly ($0.5$ = chance); w/o complexity: log PNG size regressed out. \textbf{Bold}: best; \underline{underline}: second best in the concordance block.}
\label{tab:offline}
\vspace{-0.6em}
\small
\setlength{\tabcolsep}{4pt}
\begin{tabular}{@{}lcc@{}}
\toprule
 & GenEval & OCR \\
\midrule
\multicolumn{3}{@{}l}{\textit{Within-group correlation}} \\
Absolute $S$ vs.\ complexity & $-0.75$ & $-0.74$ \\
\method $D$ vs.\ complexity & $\phantom{-}0.12$ & $\phantom{-}0.11$ \\
\method $D$ vs.\ absolute $S$ & $-0.11$ & $-0.09$ \\
\midrule
\multicolumn{3}{@{}l}{\textit{Within-group concordance with correctness} $\uparrow$} \\
Complexity (log PNG size) & \underline{0.53} & 0.50 \\
Absolute $S$ & 0.48 & \underline{0.51} \\
\rowcolor{oursrow}\method $D$ & \textbf{0.60} & \textbf{0.63} \\
\midrule
\multicolumn{3}{@{}l}{\textit{Diagnostic}} \\
\method $D$ (w/o complexity) & 0.60 & 0.63 \\
\bottomrule
\end{tabular}
\vspace{-0.8em}
\end{table}

\section{Experiments}
\label{sec:exp}

\begin{figure*}[tp]
\centering
\setlength{\qbw}{0.495\textwidth}%
\begin{tabular}[t]{@{}c@{}}
\makebox[0.333\qbw][c]{\scriptsize Base model}\makebox[0.333\qbw][c]{\scriptsize + SOLACE}\makebox[0.333\qbw][c]{\scriptsize + \method} \\[0pt]
{\small DrawBench} \\[-1pt]
\includegraphics[width=\qbw]{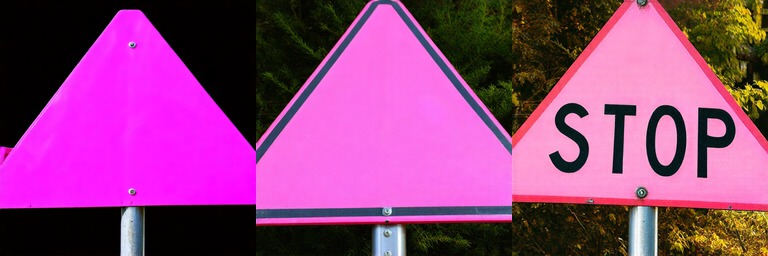} \\[-3.5pt]
\parbox[t]{\qbw}{\centering\scriptsize A triangular pink stop sign. A pink stop sign in the shape of a triangle.} \\[2.5pt]
\includegraphics[width=\qbw]{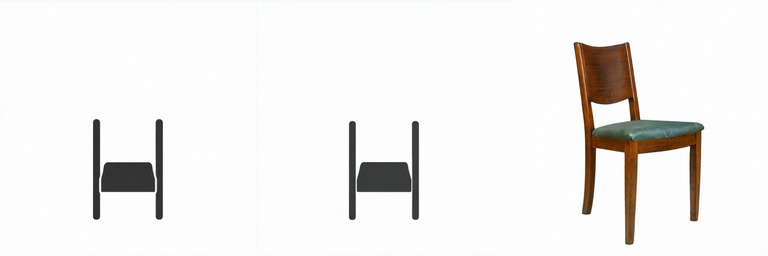} \\[-3.5pt]
\parbox[t]{\qbw}{\centering\scriptsize A separate seat for one person, typically with a back and four legs.} \\[2.5pt]
\includegraphics[width=\qbw]{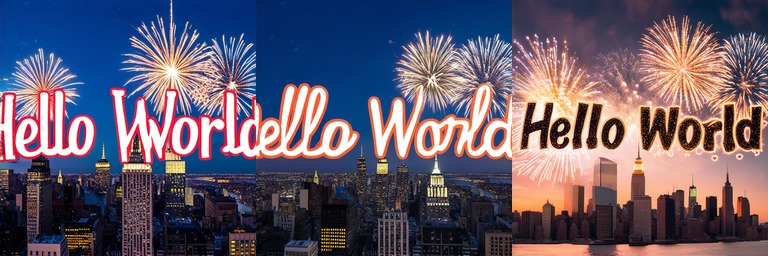} \\[-3.5pt]
\parbox[t]{\qbw}{\centering\scriptsize New York Skyline with 'Hello World' written with fireworks on the sky.} \\[2.5pt]
\end{tabular}
\hfill\begin{tabular}[t]{@{}c@{}}
\makebox[0.333\qbw][c]{\scriptsize Base model}\makebox[0.333\qbw][c]{\scriptsize + SOLACE}\makebox[0.333\qbw][c]{\scriptsize + \method} \\[0pt]
{\small GenEval} \\[-1pt]
\includegraphics[width=\qbw]{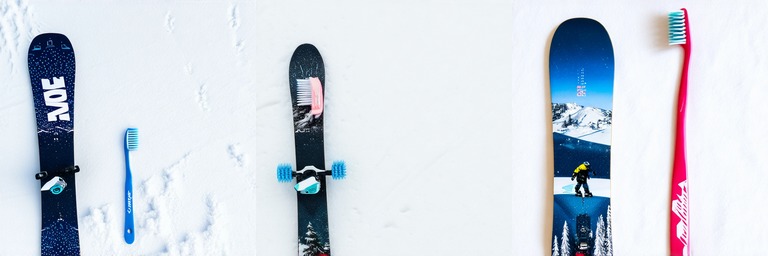} \\[-3.5pt]
\parbox[t]{\qbw}{\centering\scriptsize a photo of a toothbrush and a snowboard} \\[2.5pt]
\includegraphics[width=\qbw]{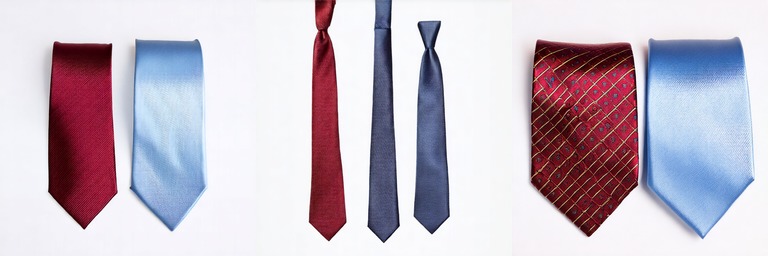} \\[-3.5pt]
\parbox[t]{\qbw}{\centering\scriptsize a photo of two ties} \\[2.5pt]
\includegraphics[width=\qbw]{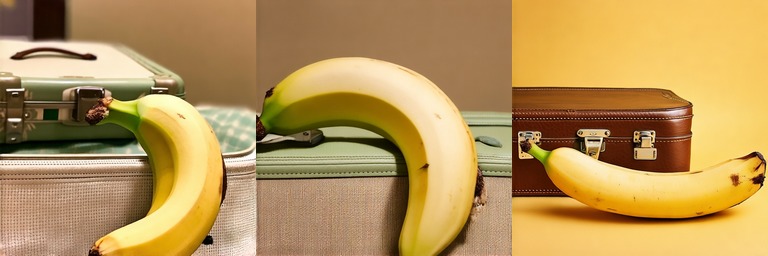} \\[-3.5pt]
\parbox[t]{\qbw}{\centering\scriptsize a photo of a suitcase left of a banana} \\[2.5pt]
\end{tabular}
\\[0pt]
\begin{tabular}[t]{@{}c@{}}
{\small OCR} \\[-1pt]
\includegraphics[width=\qbw]{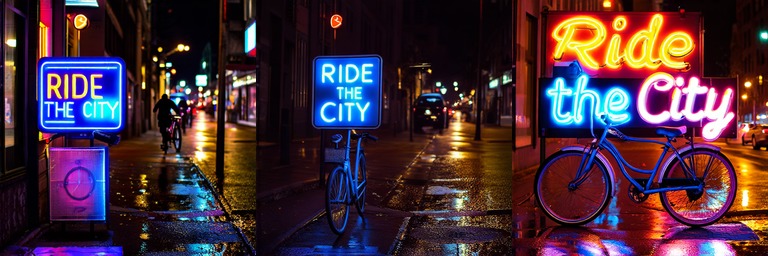} \\[-3.5pt]
\parbox[t]{\qbw}{\centering\scriptsize ``Ride the City''} \\[2.5pt]
\includegraphics[width=\qbw]{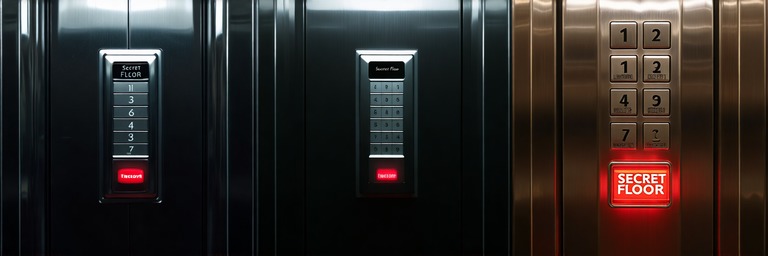} \\[-3.5pt]
\parbox[t]{\qbw}{\centering\scriptsize ``Secret Floor''} \\[2.5pt]
\includegraphics[width=\qbw]{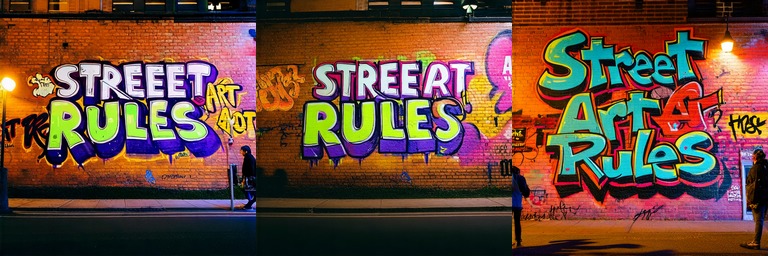} \\[-3.5pt]
\parbox[t]{\qbw}{\centering\scriptsize ``Street Art Rules''} \\[2.5pt]
\end{tabular}
\hfill\begin{tabular}[t]{@{}c@{}}
{\small Pick-a-Pic} \\[-1pt]
\includegraphics[width=\qbw]{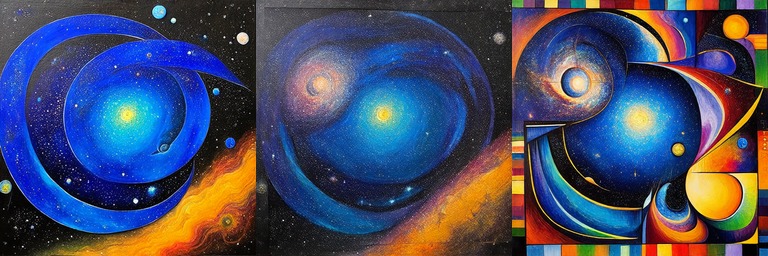} \\[-3.5pt]
\parbox[t]{\qbw}{\centering\scriptsize picasso style art of the galaxy} \\[2.5pt]
\includegraphics[width=\qbw]{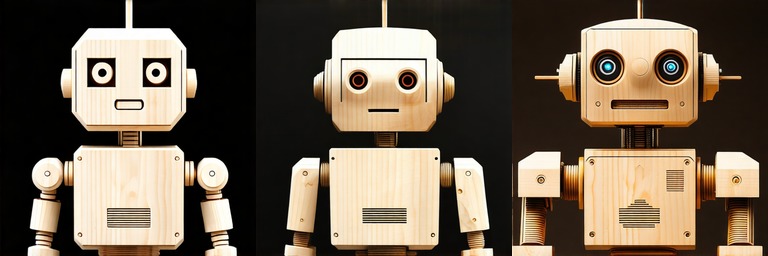} \\[-3.5pt]
\parbox[t]{\qbw}{\centering\scriptsize a futuristic robot made of wood} \\[2.5pt]
\includegraphics[width=\qbw]{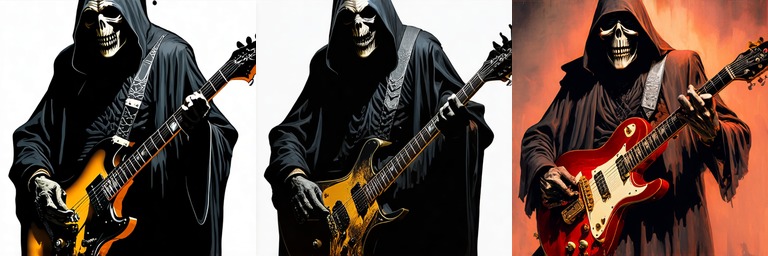} \\[-3.5pt]
\parbox[t]{\qbw}{\centering\scriptsize Grim reaper playing an electric guitar} \\[2.5pt]
\end{tabular}

\vspace{-0.3em}
\caption{\textbf{Qualitative comparison on four prompt sets.} DrawBench, GenEval, OCR and Pick-a-Pic test prompts; shared initial noise. OCR labels: text to render. Hand-picked to show differences.}
\label{fig:qual}
\vspace{-0.5em}
\end{figure*}

\begin{table*}[t]
\centering
\caption{\textbf{Post-training results.} Measured in this work; GenEval: official task-averaged; OCR: Flow-GRPO set; rest: DrawBench; SDXL, SD3.5-L: unranked. Flow-GRPO (X): public adapter for X. + Y: Base, or the adapter above if indented, post-trained with reward Y. \textbf{Bold}: best; \underline{underline}: second best per block; ties share the mark.}
\label{tab:main}
\vspace{-0.6em}
\small
\setlength{\tabcolsep}{2.1pt}
\begin{tabular}{@{}lcccccccc@{}}
\toprule
 & \multicolumn{2}{c}{Task-specific} & \multicolumn{2}{c}{\shortstack{Text--image alignment\\\& aesthetics}} & \multicolumn{4}{c}{Human preference} \\
\cmidrule(lr){2-3}\cmidrule(lr){4-5}\cmidrule(lr){6-9}
Model & GenEval$\uparrow$ & OCR$\uparrow$ & CLIPScore$\uparrow$ & Aesthetic$\uparrow$ & PickScore$\uparrow$ & HPSv2.1$\uparrow$ & ImageReward$\uparrow$ & UnifiedReward$\uparrow$ \\
\midrule
SDXL & 0.54 & 0.13 & 0.278 & 5.60 & 22.30 & 0.265 & 0.55 & 2.84 \\
SD3.5-L & 0.70 & 0.70 & 0.290 & 5.53 & 22.79 & 0.289 & 0.97 & 3.25 \\
Base & \underline{0.69} & \underline{0.63} & \underline{0.284} & \underline{5.38} & \underline{22.37} & \underline{0.283} & \underline{0.91} & \underline{3.11} \\
\;+ SOLACE & \underline{0.69} & \underline{0.63} & \textbf{0.285} & 5.35 & 22.36 & 0.276 & 0.86 & 3.09 \\
\rowcolor{oursrow}\;+ \method & \textbf{0.71} & \textbf{0.70} & \textbf{0.285} & \textbf{5.45} & \textbf{22.56} & \textbf{0.297} & \textbf{1.09} & \textbf{3.20} \\
\midrule
\;+ Flow-GRPO (GenEval) & \textbf{0.94} & \underline{0.71} & \underline{0.294} & \underline{5.33} & \underline{22.53} & \underline{0.280} & \underline{1.12} & \textbf{3.27} \\
\quad + SOLACE & \underline{0.92} & 0.68 & 0.292 & \underline{5.33} & 22.47 & 0.270 & 1.00 & \underline{3.25} \\
\rowcolor{oursrow}\quad + \mbox{\method} & \underline{0.92} & \textbf{0.73} & \textbf{0.301} & \textbf{5.52} & \textbf{22.73} & \textbf{0.305} & \textbf{1.28} & 3.20 \\
\midrule
\;+ Flow-GRPO (OCR) & \underline{0.70} & \textbf{0.92} & \underline{0.290} & \textbf{5.31} & \underline{22.45} & \underline{0.284} & \underline{0.99} & \textbf{3.16} \\
\quad + SOLACE & 0.69 & \underline{0.89} & 0.289 & \underline{5.30} & 22.38 & 0.277 & 0.92 & \underline{3.13} \\
\rowcolor{oursrow}\quad + \mbox{\method} & \textbf{0.74} & \textbf{0.92} & \textbf{0.293} & 5.24 & \textbf{22.55} & \textbf{0.293} & \textbf{1.13} & 3.11 \\
\midrule
\;+ Flow-GRPO (PickScore) & \textbf{0.75} & \underline{0.73} & 0.280 & \underline{5.91} & \underline{23.47} & \underline{0.321} & \underline{1.31} & \underline{3.41} \\
\quad + SOLACE & 0.72 & 0.68 & \textbf{0.287} & 5.54 & 22.70 & 0.281 & 1.04 & 3.25 \\
\rowcolor{oursrow}\quad + \mbox{\method} & \underline{0.74} & \textbf{0.76} & \underline{0.281} & \textbf{5.97} & \textbf{23.55} & \textbf{0.322} & \textbf{1.36} & \textbf{3.43} \\
\bottomrule
\end{tabular}
\vspace{-0.5em}
\end{table*}

\subsection{Setup}
\label{sec:setup}
\paragraph{Training.}\looseness=-1 We post-train the base model, Stable Diffusion 3.5 Medium~\cite{esser2024sd3}, with Flow-GRPO~\cite{liu2025flowgrpo}, using the intrinsic reward of \cref{eq:reward} as the only reward, on the OCR training prompts of Flow-GRPO with group size $G\!=\!16$ and rollouts at CFG $4.5$. The reward uses $K\!=\!8$ antithetic probes at $|\mathcal{T}|\!=\!5$ sampler steps. We fine-tune LoRA~\cite{hu2022lora} adapters with rank $32$ and $\alpha\!=\!64$ for $2{,}000$ iterations, using seed $42$ and $8$ H100 GPUs with $8$ samples per GPU. Apart from the GPU type, these settings coincide with those reported for SOLACE~\cite{kim2026solace}, so the reported settings of the two methods differ only in the reward. We use AdamW with a constant learning rate of $3\times10^{-4}$ and a KL weight of $\beta\!=\!0.04$. Rollouts use $10$ SDE steps with noise scale $0.7$. As in Flow-GRPO, the last step is not trained, and losses are applied to the last $\lceil\rho\cdot 9\rceil\!=\!6$ of the remaining $9$ steps with $\rho\!=\!0.6$. Evaluation uses exponential-moving-average weights with decay $0.9$.

\paragraph{Evaluation.} We evaluate with external metrics that the reward never sees, in four categories that are standard for intrinsic-reward and external-reward T2I post-training~\cite{liu2025flowgrpo,kim2026solace}. Unless stated otherwise, we generate images at $512\!\times\!512$ resolution with $40$ inference steps and CFG $7.0$. SDXL~\cite{podell2024sdxl} and Stable Diffusion 3.5 Large (SD3.5-L)~\cite{esser2024sd3} use their recommended settings, and SOLACE is evaluated from its public checkpoint. The base model and all models post-trained from it share the per-sample initial noise.
(1) \emph{Compositional generation}: GenEval~\cite{ghosh2023geneval} with the official detector-based pipeline, $553$ prompts $\times$ $4$ samples.
(2) \emph{Visual text rendering}: the $1{,}018$ OCR test prompts of Flow-GRPO~\cite{liu2025flowgrpo}, scored as $\max(0,1-N_e/N_{\text{ref}})$ ($N_e$: edit distance; $N_{\text{ref}}$: target-text length) with PaddleOCR~\cite{cui2025paddleocr}.
(3) \emph{Human preference}: PickScore~\cite{kirstain2023pickscore}, HPSv2.1~\cite{wu2023hpsv2}, ImageReward~\cite{xu2023imagereward} and UnifiedReward~\cite{wang2025unifiedreward} on DrawBench~\cite{saharia2022imagen}.
(4) \emph{Text--image alignment and aesthetics}: CLIPScore~\cite{hessel2021clipscore,radford2021clip} and the LAION aesthetic score~\cite{schuhmann2022laion} on DrawBench.
Qualitative figures also use test prompts of Pick-a-Pic~\cite{kirstain2023pickscore}. Offline-analysis, training and evaluation details are in the supplementary material.

\begin{table}[t]
\centering
\caption{\textbf{Comparison with absolute self-confidence.} Measured in this work; GenEval: official task-averaged; OCR: Flow-GRPO set; rest: DrawBench. CLIP: CLIPScore; Aes.: aesthetic; Pick: PickScore; HPS: HPSv2.1; IR: ImageReward; Iter.: \method training iteration; --: not applicable. \textbf{Bold}: best; \underline{underline}: second best per column; ties share the mark.}
\label{tab:robust}
\vspace{-0.6em}
\small
\setlength{\tabcolsep}{3.5pt}
\begin{tabular}{@{}lccccccc@{}}
\toprule
Iter. & GenEval$\uparrow$ & OCR$\uparrow$ & CLIP$\uparrow$ & Aes.$\uparrow$ & Pick$\uparrow$ & HPS$\uparrow$ & IR$\uparrow$ \\
\midrule
\multicolumn{8}{@{}l}{\textit{Base}} \\
-- & 0.687 & 0.626 & 0.284 & 5.38 & 22.37 & 0.283 & 0.91 \\
\multicolumn{8}{@{}l}{\textit{+ SOLACE}} \\
-- & 0.692 & 0.626 & 0.285 & 5.35 & 22.36 & 0.276 & 0.86 \\
\multicolumn{8}{@{}l}{\textit{+ \method}} \\
\rowcolor{oursrow}1{,}040 & \textbf{0.715} & 0.701 & \textbf{0.288} & \underline{5.41} & 22.54 & 0.294 & \underline{1.07} \\
\rowcolor{oursrow}1{,}520 & 0.703 & \textbf{0.722} & \underline{0.286} & \textbf{5.45} & \textbf{22.59} & \underline{0.295} & \underline{1.07} \\
\rowcolor{oursrow}2{,}000 & \underline{0.706} & \underline{0.704} & 0.285 & \textbf{5.45} & \underline{22.56} & \textbf{0.297} & \textbf{1.09} \\
\bottomrule
\end{tabular}
\vspace{-0.8em}
\end{table}

\subsection{Comparison with the base model and absolute self-confidence}
\label{sec:results}

\method improves all eight metrics over the base model (\cref{tab:main}), including the human-preference scores on which SOLACE~\cite{kim2026solace} reports little or no gain. The CLIPScore gain is small ($0.284$ to $0.285$). Under the same evaluation protocol, \method also improves over SOLACE, the absolute self-confidence baseline (\cref{tab:robust}), on OCR score ($0.626$ to $0.704$), the aesthetic score and the three preference scores. It is equal on CLIPScore ($0.285$). On GenEval, \method is higher than SOLACE ($0.706$ against $0.692$). A policy trained on $S$ with our recipe and only the reward changed, unlike the public SOLACE checkpoint, falls below the base model on every metric, with GenEval $0.474$ and OCR $0.446$. Its samples are simpler, as the within-group preference of $S$ for simple samples (\cref{tab:offline}) predicts. Their mean PNG size is $213$\,KiB on GenEval and $203$\,KiB on OCR, against $334$ and $377$\,KiB for the base model and $374$ and $418$\,KiB for \method. Full results of this policy and published values for the baselines and the Flow-GRPO adapters are in the supplementary material.

\subsection{Comparison with classifier-free guidance}
\label{sec:guidance}
Stronger guidance of the base model does not reproduce the text-rendering, aesthetic and preference gains of \method, although post-training on the implicit-classifier score $D$ might act like a larger CFG scale. We evaluate the base model at CFG scales from $4.5$ to $13.0$ and compare it with \method at the fixed CFG scale of $7.0$, giving the base model the best of five scales per metric, selected on the same test prompts (\cref{tab:cfg}). Stronger guidance raises OCR score only up to $0.644$, whereas \method reaches $0.704$. On GenEval, the score is $0.690$ at the best base scale ($9.0$) and $0.706$ for \method, so \method is comparable to tuned guidance on this benchmark. Aesthetic, PickScore, HPSv2.1 and ImageReward of the base model peak between CFG $4.5$ and $9.0$ and then decrease, whereas \method is above every scale on all four.

\begin{table}[t]
\centering
\caption{\textbf{Base model across CFG scales \vs \method at CFG 7.0.} Measured in this work; GenEval: official task-averaged; OCR: Flow-GRPO set; rest: DrawBench. CLIP: CLIPScore; Aes.: aesthetic; Pick: PickScore; HPS: HPSv2.1; IR: ImageReward. \textbf{Bold}: best; \underline{underline}: second best per column; ties share the mark.}
\label{tab:cfg}
\vspace{-0.6em}
\footnotesize
\setlength{\tabcolsep}{2.6pt}
\begin{tabular}{@{}lccccccc@{}}
\toprule
CFG & GenEval$\uparrow$ & OCR$\uparrow$ & CLIP$\uparrow$ & Aes.$\uparrow$ & Pick$\uparrow$ & HPS$\uparrow$ & IR$\uparrow$ \\
\midrule
\multicolumn{8}{@{}l}{\textit{Base}} \\
4.5 & 0.671 & 0.577 & 0.283 & \underline{5.38} & 22.35 & 0.280 & 0.83 \\
7.0 & 0.687 & 0.626 & \underline{0.284} & \underline{5.38} & \underline{22.37} & \underline{0.283} & 0.91 \\
9.0 & \underline{0.690} & 0.643 & \underline{0.284} & 5.37 & 22.32 & \underline{0.283} & \underline{0.93} \\
11.0 & 0.684 & \underline{0.644} & \textbf{0.285} & 5.33 & 22.22 & 0.280 & 0.90 \\
13.0 & 0.655 & \underline{0.644} & 0.283 & 5.28 & 22.08 & 0.275 & 0.84 \\
\midrule
\rowcolor{oursrow}\method, 7.0 & \textbf{0.706} & \textbf{0.704} & \textbf{0.285} & \textbf{5.45} & \textbf{22.56} & \textbf{0.297} & \textbf{1.09} \\
\bottomrule
\end{tabular}
\vspace{-0.8em}
\end{table}

\subsection{Continued training of Flow-GRPO adapters}
\label{sec:adapters}
In the Flow-GRPO blocks of \cref{tab:main}, continuing each public adapter with \method as the only reward and the same training settings raises PickScore and ImageReward for all three adapters. The score of the external reward the adapter was trained on rises for PickScore, is unchanged for OCR, and falls for GenEval, from $0.94$ to $0.92$. The other costs are lower UnifiedReward for the GenEval and OCR adapters, a lower aesthetic score for the OCR adapter, and a lower GenEval score for the PickScore adapter ($0.75$ to $0.74$). All other metrics are higher than for the adapter. Continuing with SOLACE instead, in its authors' public checkpoints, lowers OCR and PickScore for all three adapters, and \method is above it on OCR, PickScore, HPSv2.1 and ImageReward for all three.

\subsection{Qualitative comparison}
\label{sec:qual}
\method renders requested text that the base model garbles or omits, \eg, ``Abandon All Hope'' (\cref{fig:teaser}) and ``Street Art Rules'' (\cref{fig:qual}), and satisfies constraints that the base model misses, \eg, an elephant below a surfboard (\cref{fig:teaser}). The prompts in these figures are selected to show such differences. With an OCR score of $0.70$ (\cref{tab:main}), \method still misrenders text on many prompts. All figures show one sample per model and prompt.

\begin{table}[t]
\centering
\caption{\textbf{Offline ablation of the reward design.} Base-model test-prompt rollouts at CFG 7.0. Concordance: fraction of same-prompt pairs ranked correctly ($0.5$ = chance). $c'$: another prompt of the same benchmark. \textbf{Bold}: best; \underline{underline}: second best per column within each block, ranked before rounding.}
\label{tab:ablation}
\vspace{-0.6em}
\small
\setlength{\tabcolsep}{4pt}
\begin{tabular}{@{}lcc@{}}
\toprule
 & GenEval$\uparrow$ & OCR$\uparrow$ \\
\midrule
\multicolumn{3}{@{}l}{\textit{Reward composition}} \\
Absolute $S=-\log\mathrm{MSE}(c)$ & 0.48 & 0.51 \\
Unconditional, $-\log\mathrm{MSE}(\varnothing)$ & 0.48 & 0.50 \\
$S+D$ & \underline{0.57} & \underline{0.60} \\
\rowcolor{oursrow}$D$ (Ours) & \textbf{0.60} & \textbf{0.63} \\
\midrule
\multicolumn{3}{@{}l}{\textit{Reference condition}} \\
Mismatched prompt $c'$ & \textbf{0.62} & \underline{0.61} \\
\rowcolor{oursrow}Empty prompt $\varnothing$ (Ours) & \underline{0.60} & \textbf{0.63} \\
\midrule
\multicolumn{3}{@{}l}{\textit{Normalization and weighting}} \\
No per-timestep z-score & \textbf{0.62} & \underline{0.65} \\
z-score, two lowest-noise $t$ dropped & \underline{0.62} & \textbf{0.65} \\
$w(t)=t(1-t)$ & 0.60 & 0.62 \\
\rowcolor{oursrow}z-score, $w(t)=1$ (Ours) & 0.60 & 0.63 \\
\bottomrule
\end{tabular}
\vspace{-0.8em}
\end{table}

\subsection{Reward-design ablation}
\label{sec:ablation}
\looseness=-1 The contrast with a reference prediction carries the prompt-specific information, while the choice of reference condition and the $t(1-t)$ weighting change concordance only slightly (\cref{tab:ablation}, one component varied at a time on the same probes, without training). \emph{Composition:} absolute and unconditional confidence stay near chance, and adding $S$ back to $D$ reintroduces the complexity bias (correlation with log PNG size $-0.47$ on GenEval, $-0.46$ on OCR) and lowers concordance on both benchmarks. \emph{Reference condition:} a mismatched prompt, a fixed random prompt of the same benchmark, works about as well as the empty prompt, which needs no extra prompt sampling. \emph{Normalization:} removing the per-timestep z-score, or keeping it but dropping the two lowest-noise timesteps, raises concordance to $0.62$ on GenEval and $0.65$ on OCR. The z-score gives the lowest-noise timestep, whose log-ratio alone is near chance ($0.52$ and $0.51$), the same weight as the high-noise timesteps that carry the signal. We keep the per-timestep z-score of SOLACE~\cite{kim2026solace} so that both rewards share one recipe. Without it, $S$ stays at chance ($0.48$ and $0.50$), so the comparison of the two rewards does not depend on this choice.

\subsection{Reward diagnostics}
\label{sec:diagnostics}
\paragraph{Unconditional prediction after training.}
Post-training with \method does not worsen the empty-prompt prediction, although a policy could raise $D$ that way without improving its prompt-conditioned one. We re-score fixed base-model rollouts with each post-trained model and identical probes (supplementary material). Compared with the base model, the \method model has a slightly lower unconditional log error ($-0.9869$ against $-0.9834$) and an unchanged conditional log error ($-1.0036$ against $-1.0035$). SOLACE also has a lower unconditional log error.

\paragraph{Advantage-sign disagreement with absolute self-confidence.}
During \method training, the group-normalized advantages under \method and under absolute self-confidence have opposite signs on $49\%$ of samples, the rate expected for uncorrelated rewards and consistent with their near-zero offline correlation (\cref{tab:offline}). The advantage depends on a reward only through its deviation from the group mean, so replacing absolute self-confidence changes the update.

\section{Conclusion}
\label{sec:conclusion}
We propose \method, an intrinsic reward for text-to-image post-training that measures how much the prompt helps the generator denoise its own sample. Absolute self-confidence, the generator's own denoising accuracy, is dominated by image complexity within same-prompt groups. As the only GRPO reward, \method improves text rendering, aesthetic and human-preference scores over the base model and a prior absolute self-confidence reward, gains that stronger classifier-free guidance does not reproduce. Its null-referenced log-ratio of the unconditional to the conditional error cancels the prompt-independent difficulty that scales both. Self-denoising thus signals prompt fidelity through the contrast between the conditional and unconditional predictions, not through absolute denoising accuracy.

\paragraph{Limitations.} We evaluate \method on a single base model, so applying it to other T2I models would test how general its gains are. \method rewards prompt fidelity, so optimizing a specific preference objective still requires a reward for that objective. \method relies on an unconditional branch, so it requires a model trained with condition dropout, which is standard for CFG-capable T2I models. Its reward needs an unconditional forward pass alongside each conditional one, so each probe costs two network evaluations. Like other post-training methods, \method can strengthen harmful content that a prompt requests.

\section*{LLM Usage Statement}
Large language models (Claude, Anthropic, used through Claude Code; and Codex, OpenAI) assisted in writing and editing the text of this paper and its supplementary material, including checking the text against the venue's formatting and style guidelines. They also wrote parts of the training, evaluation, and analysis code. The authors reviewed all edits and take full responsibility for the content.

{
    \small
    \bibliographystyle{ieeenat_fullname}
    \bibliography{main}
}

\end{document}